\documentclass[]{kwaivideo/kwaivideo}

\usepackage{microtype}
\usepackage{graphicx}
\usepackage{subcaption}
\usepackage{booktabs} 
\usepackage{csquotes}
\usepackage{afterpage}
\usepackage{xcolor}

\usepackage{amsmath}
\usepackage{amssymb}
\usepackage{mathtools}
\usepackage{amsthm}
\usepackage{xspace}
\usepackage{colortbl}
\usepackage{multirow}
\usepackage{enumitem}
\usepackage{wrapfig}
\usepackage{nicefrac}
\usepackage[font=small,labelfont=bf,justification=centering]{caption}

\usepackage{tabularx}
\usepackage{array}
\usepackage{siunitx} 

\newcolumntype{Y}{>{\raggedright\arraybackslash}X}

\definecolor{highlightrow}{HTML}{EAF1FB}

\title{LPM: Industrial-Scale Generative Video Restoration}
\headertext{LPM: Industrial-Scale Generative Video Restoration}

\author[]{LPM Team}
\affiliation[]{Kuaishou Technology}
\affiliation[]{Tsinghua University}

\abstract{We present the \textbf{Large Processing Model (LPM)}, a diffusion-based generative framework for photorealistic video restoration under complex, in-the-wild degradations. To our knowledge, LPM is the \textbf{first generative video restoration model deployed at industrial scale}. LPM addresses the diverse degradations in user-generated content (UGC) through a unified system encompassing large-scale data engineering, foundation-model training, and efficient inference. Its enhanced architecture, progressive training strategy, and temporal-pyramid inference mechanism jointly enable high-fidelity, temporally consistent restoration of arbitrarily long videos across the broad content distribution encountered on UGC platforms. LPM has been deployed in production at Kuaishou, where videos processed by the model account for approximately \textbf{45\% of total viewing time}, delivering consistent improvements across key quality-of-experience metrics. Beyond perceptual enhancement, LPM delivers substantial system-level benefits: at comparable perceptual quality, it reduces bitrate by \textbf{20\%} relative to Kuaishou's in-house codec, yielding annual bandwidth cost savings on the order of \textbf{hundreds of millions}. Its low serving cost also enables integration into products such as \textbf{Kling}, demonstrating that generative restoration can be practical, scalable, and cost-effective for large-scale video processing.}

\metadata[Date]{October 1, 2025}

\begin{document}

\maketitle

\section{Introduction}
\label{section:intro}

Video has become a primary medium for communication, content creation, and information consumption worldwide~\cite{cloudflare_radar_yir_2025, ericsson_mobility_2025_video_share}. However, a substantial fraction of real-world video is captured and delivered under imperfect conditions. In particular, user-generated content (UGC) routinely suffers from noise, blur, compression artifacts, and other distortions caused by heterogeneous capture devices, repeated transcoding, and constrained transmission bandwidth~\cite{youtube_ugc}. These degradations directly impair the viewing experience at platform scale. Robust restoration of in-the-wild video is therefore important not only for human viewers, but also for multimodal systems~\cite{achiam2023gpt, team2023gemini, liu2024deepseek} that rely on high-quality visual signals for perception and reasoning.

Despite substantial progress in image and video restoration, conventional methods often struggle to generalize to complex, in-the-wild degradations~\cite{zhu2024cpga, mo2024oapt, zeng2026shiftlut, bao2025plug}.
Diffusion-based generative models~\cite{ho2020ddpm, rombach2022highresolution} have further advanced restoration by formulating it as conditional generation from degraded inputs to high-fidelity outputs. By leveraging generative priors learned from large-scale data, these models can recover realistic textures and structures that are difficult to reconstruct with conventional restoration methods. This paradigm has achieved strong results in image restoration~\cite{wang2024exploiting, lin2023diffbir, chen2024cassr, yu2024scaling, wu2024seesr, yu2025zipir, wang2026coloring, bai2026tuning} and has increasingly been extended to video enhancement~\cite{zhou2024upscale, yang2024motion, wang2025seedvr}. Video restoration architectures have evolved from augmenting image backbones with temporal modules~\cite{zhou2024upscale} to Diffusion Transformers (DiTs) that exploit pre-trained spatiotemporal priors~\cite{peebles2023scalable, wang2025seedvr, li2025diffvsr, xie2025star}. These advances improve structural integrity and temporal coherence, while recent distillation techniques~\cite{zhuang2025flashvsr, bai2026instantvir} have further reduced inference cost.

Despite this progress, existing methods do not yet meet the combined requirements of industrial-scale video restoration. Image restoration models can generate fine spatial details, but lack explicit temporal modeling; when applied independently to individual frames, they often produce flickering and inconsistent textures. Video models initialized from text-to-video (T2V) backbones improve temporal coherence, but joint spatiotemporal modeling is computationally expensive and can compromise per-frame fidelity, often resulting in over-smoothed outputs. Moreover, most existing methods are trained and evaluated on short clips with fixed temporal windows. Directly applying them to long-form UGC can lead to discontinuities between windows, progressive appearance drift, and instability around scene transitions. Achieving both fine-grained spatial fidelity and long-term temporal consistency across arbitrary-length, in-the-wild videos therefore remains an open challenge.

We address this challenge with the \textbf{Large Processing Model (LPM)}, a diffusion-based generative framework for high-fidelity video restoration across diverse real-world content and degradation patterns. LPM unifies large-scale data engineering, progressive generative training, and efficient inference in a single restoration system. It is designed to preserve the detail-generation capability of image diffusion models while introducing effective temporal modeling for stable video restoration. LPM has been deployed as a server-side automatic restoration service in Kuaishou. It currently processes videos accounting for approximately 45\% of the platform's total viewing time and consistently improves key online quality-of-experience (QoE) metrics. To our knowledge, LPM is the first generative video restoration model deployed at industrial scale.

LPM consists of two stages: \textit{LPM-Image} and \textit{LPM-Video}. LPM-Image establishes a strong spatial restoration prior through a large-scale data pipeline covering diverse content and degradation distributions. It is trained progressively, first learning coarse-grained structure restoration and then fine-grained texture generation. LPM-Video extends this image prior to the temporal domain through efficient temporal alignment and fusion, preserving the spatial fidelity of LPM-Image while improving frame-to-frame consistency. To support long-form content, we further develop a temporal-pyramid inference framework that combines long-range guidance with short-range anchors. This design mitigates error accumulation and appearance drift, enabling temporally stable restoration of videos with arbitrary duration.

Our main contributions are summarized as follows:
\begin{itemize}
    \item \textbf{A progressive image-to-video restoration framework.}
    We introduce a progressive training paradigm that first learns conditional image restoration and subsequently extends the learned spatial prior to video. In contrast to directly fine-tuning a T2V backbone, this design better preserves fine-grained texture generation while reducing the blurring and over-smoothing commonly observed in video restoration. LPM further replaces the conventional 3D-VAE design with an efficient 2D-VAE and factorized 2D+1D DiT architecture, substantially improving training and inference efficiency.

    \item \textbf{Temporally consistent restoration of arbitrary-length videos.}
    LPM is designed without fixed temporal positional encoding, allowing it to generalize beyond the temporal windows observed during training. We further introduce temporal-pyramid inference, which integrates long-range guidance with short-range anchor frames to maintain temporal coherence, reduce cross-window discontinuities, and prevent progressive drift. Together, these designs enable stable restoration of videos with arbitrary duration.

    \item \textbf{Industrial-scale deployment and system-level efficiency.}
    LPM achieves state-of-the-art restoration quality in systematic evaluations and has been deployed at scale in Kuaishou's production video pipeline. It currently serves videos representing approximately {45\% of total viewing time} and delivers consistent improvements across key online QoE metrics. When integrated with Kuaishou's encoding and transmission pipeline, LPM reduces bitrate by more than {20\%} at comparable perceptual quality relative to the in-house codec, demonstrating the practical and economic value of generative video restoration at scale.
\end{itemize}

\section{Approach}

\subsection{Data}

\begin{figure}[tbp]
    \centering
    \includegraphics[width=0.88\textwidth]{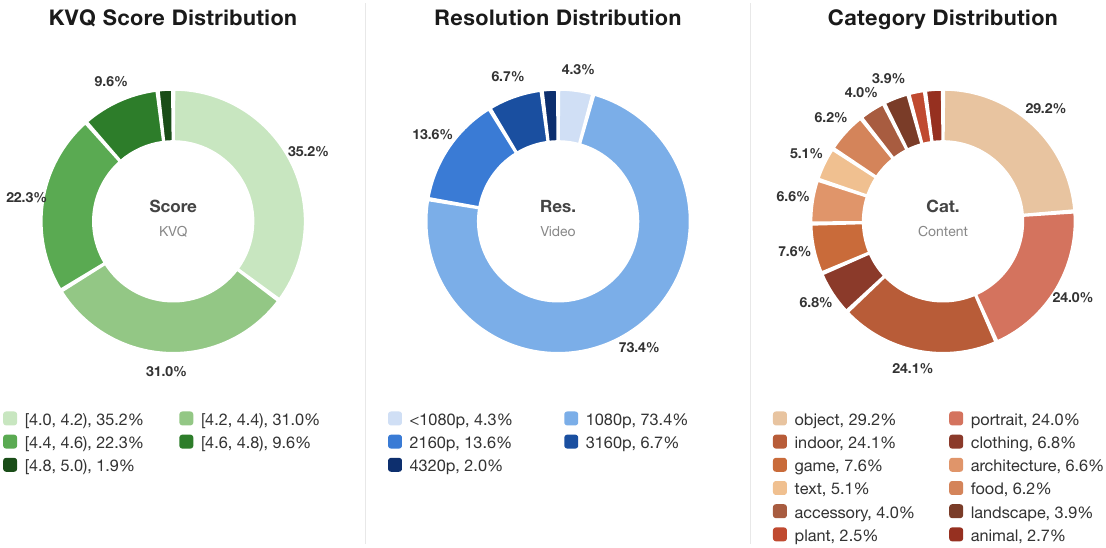}
    \caption{The KVQ scores, resolutions, and categories distributions of our Kwai UltraVision Dataset.}
    \label{fig:data_analy_fig}
\end{figure}

Large-scale, high-quality training data is fundamental to robust generative restoration. Although the text-to-image (T2I) and text-to-video (T2V) communities have benefited from a growing collection of large-scale datasets~\cite{kong2024hunyuanvideo, wan2025wan, zheng2025open}, substantially fewer data resources satisfy the stringent quality requirements of restoration and enhancement. In particular, large restoration models require clean visual targets with both high perceptual quality and abundant fine-grained detail; deficiencies in either dimension directly limit the quality of the learned generative prior. We therefore construct the \textbf{Kwai UltraVision Dataset} around two primary criteria: (1) ultra-high visual quality and (2) rich texture complexity. The resulting dataset contains visual samples at the billion scale, covers $12$ content categories, and achieves an average Kuaishou Visual Quality (KVQ) score of $4.31$ on a $[0,5]$ scale. Fig.~\ref{fig:data_analy_fig} summarizes its distributions in visual quality, resolution, and semantic category.

\subsubsection{Dataset for Large Processing Models}

T2I and T2V datasets typically prioritize semantic coverage and text--visual alignment, with the strictest quality filtering often reserved for fine-tuning. Generative restoration has different data requirements: it relies on clean visual targets to synthesize degraded inputs and learn the reconstruction of fine-grained artifacts and high-frequency details. High visual quality is therefore essential throughout training. However, existing datasets contain relatively few samples that combine exceptional visual fidelity with rich textures. We address this gap by curating data along two dimensions: visual quality and texture complexity.

\begin{itemize}
    \item \textbf{Ultra-High Visual Quality.}
    The quality of clean targets directly determines the upper bound of restoration fidelity. Targets containing blur, noise, compression artifacts, or exposure defects may cause the model to reproduce these imperfections. We therefore use Kuaishou Visual Quality (KVQ)~\cite{lu2024kvq, qu2025kvq, yuan2024ptm}, a no-reference metric that evaluates perceptual factors such as blur, noise, sharpness, and blocking artifacts. KVQ scores visual quality on a five-point scale: a score of \(4.0\) indicates clear subjects without readily perceptible artifacts, while \(4.5\) additionally requires rich textures, accurate colors, and well-preserved highlight and shadow details. We retain samples with KVQ scores above \(4.0\) for large-scale training and raise the threshold to \(4.7\) for high-quality fine-tuning. After automated filtering and manual inspection, approximately \(4.08\text{\textperthousand}\) of the candidate samples are retained for fine-tuning. The complete dataset has an average KVQ score of \(4.31\).

    \item \textbf{Rich Texture Complexity.}
    Fine structures such as foliage, animal fur, and skin texture are essential to perceptual quality. Training on samples with limited high-frequency content weakens the model's ability to reconstruct such details and often leads to over-smoothed outputs. We therefore introduce Kuaishou Texture Quality (KTQ), which measures texture complexity in both the spatial and frequency domains. The spatial component applies a Laplacian operator and measures the proportion of responsive pixels, while the frequency component uses the Discrete Cosine Transform (DCT) to compute the energy ratio in high-frequency bands. Together, they quantify the richness of fine-grained textures in each sample.
\end{itemize}

Beyond visual quality and texture complexity, we balance semantic categories and scene coverage during data construction. The resulting Kwai UltraVision Dataset combines billion-scale data, high perceptual quality, rich textures, and diverse content, providing a strong foundation for training large generative restoration models.

\subsubsection{Image Data Pipeline}

We collect raw image data from a large-scale mixture of internal sources and publicly available datasets. Our goal is to construct a training corpus that combines high perceptual quality, rich texture complexity, and broad semantic coverage for large generative restoration models. As illustrated in Fig.~\ref{fig:data_process_pipeline}, the image data pipeline consists of four stages:

\begin{figure}[tbp]
    \centering
    \includegraphics[width=0.95\linewidth]{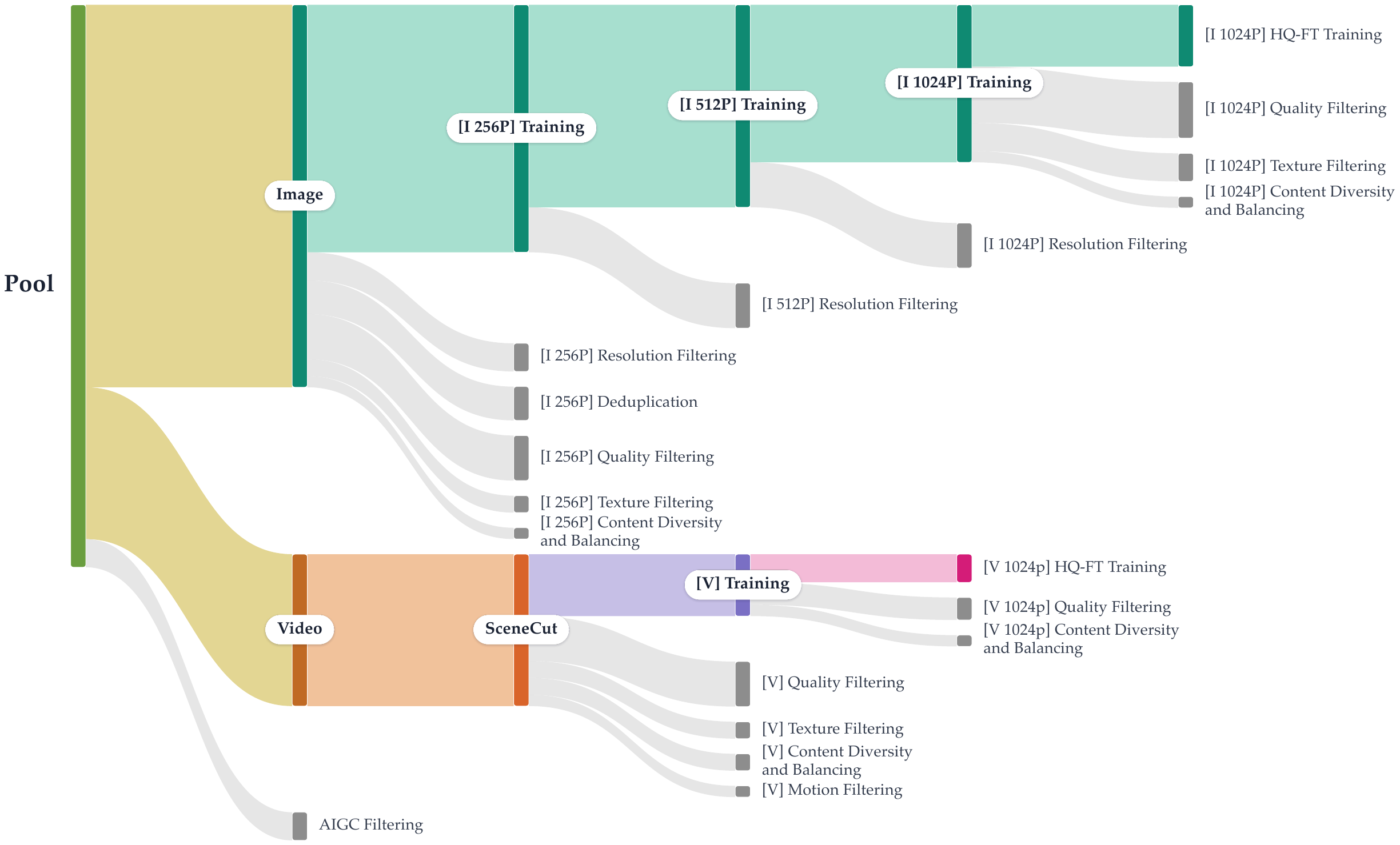}
    \caption{Overview of the data construction pipeline.}
    \label{fig:data_process_pipeline}
\end{figure}

\begin{itemize}
    \item \textbf{Quality Filtering.}
    We use KVQ to assess the perceptual quality of each candidate image and discard samples with scores below $4.0$. The filtering threshold is adapted to the requirements of each training stage: images with KVQ scores above $4.0$ are used for large-scale pre-training, whereas a substantially more selective subset with scores above $4.7$ is reserved for high-fidelity fine-tuning. We additionally remove AI-generated content (AIGC) from the training corpus. Compared with natural photographs, many generated images exhibit overly smooth surfaces, simplified local statistics, or insufficient stochastic high-frequency detail. Mixing such data with natural imagery may introduce an undesirable distribution shift and bias the learned prior toward synthetic texture patterns. Excluding AIGC therefore helps the model learn authentic image statistics and improves its generalization to real-world inputs.

    \item \textbf{Resolution Grading.}
    Our training curriculum comprises three stages: T2I base training, restoration pre-training, and high-fidelity fine-tuning. To support progressive learning across spatial scales, we organize the data into three resolution tiers of $256$, $512$, and $1024$. This coarse-to-fine curriculum stabilizes optimization at high resolutions and progressively improves the model's ability to capture both global structures and fine-scale details.

    \item \textbf{Texture Filtering.}
    Generative restoration requires the model to reconstruct realistic textures in regions where high-frequency information has been degraded or removed. Training predominantly on flat or low-complexity content provides limited supervision for this capability and can encourage over-smoothed predictions. We therefore use KTQ to measure texture richness and filter out samples with insufficient texture complexity, retaining images that contain abundant high-frequency structures and fine-grained details.

    \item \textbf{Content Diversity and Balancing.}
    Broad semantic coverage is essential for robust generalization across real-world scenes. We first extract CLIP embeddings and remove near-duplicate images according to their cosine similarity. We then use CLIP-based zero-shot classification to assign the remaining samples to semantic categories and perform category-aware resampling to improve distributional balance.
\end{itemize}

Together, these stages produce three stage-specific data tiers for T2I base training, restoration pre-training, and high-fidelity fine-tuning, respectively. Each tier is constructed with quality thresholds, resolution distributions, texture requirements, and category compositions tailored to its corresponding training objective.

\subsubsection{Video Data Pipeline}

Video restoration introduces the additional challenge of modeling temporal dynamics. Inadequate temporal modeling can lead to jitter, flickering, inconsistent textures, and motion artifacts. Beyond the image-level criteria of perceptual quality, resolution, texture complexity, and content diversity, we therefore impose video-specific constraints on temporal quality, shot continuity, and motion distribution.
Fig.~\ref{fig:data_process_pipeline} summarizes the complete data construction pipeline, from raw data collection to image-level filtering and video-specific temporal curation.

\begin{itemize}
    \item \textbf{Temporal Quality Filtering.}
    LPM-Video is trained on temporally continuous clips with stage-specific lengths. A single severely degraded frame may contaminate temporal feature aggregation and provide unreliable supervision to neighboring frames. We therefore require every frame within a retained training clip to have a KVQ score above $4.0$. We further construct stage-specific video data tiers whose clip lengths match the temporal configurations used at the corresponding training stages.

    \item \textbf{Scene-Cut Detection.}
    Learning effective temporal priors requires coherent motion and appearance within each training sample. Scene cuts create abrupt discontinuities in both visual content and motion, violating this assumption and potentially interfering with temporal alignment and feature aggregation. We apply an internal scene-cut detector to identify shot boundaries and segment each source video accordingly, ensuring that every training clip lies within a single temporally coherent shot.

    \item \textbf{Motion Balancing.}
    A diverse motion distribution is important for learning robust temporal dynamics. We estimate dense optical flow between adjacent frames using an internal optical-flow model and characterize each clip by its average flow magnitude. Based on this statistic, clips are partitioned into static, low-motion, and high-motion groups. We then rebalance these groups during data construction. This strategy exposes the model to a broad range of motion patterns, helping it preserve sharp details under large motion while maintaining stable outputs in nearly static scenes.
\end{itemize}

\subsection{Image-Based Model}

This section introduces LPM-Image, including its architecture and three-stage training pipeline: large-scale pre-training, high-quality fine-tuning, and fidelity refinement.

\subsubsection{Architecture}

\paragraph{\textbf{Improved Diffusion Transformer.}}
LPM-Image builds on the Diffusion Transformer (DiT) architecture~\cite{peebles2023scalable}, with several modifications designed to improve reconstruction fidelity, model capacity, and optimization efficiency. For latent-space compression, we develop \textbf{LPM-VAE}, a variational autoencoder (VAE)~\cite{kingma2013auto} with a moderate compression ratio and high reconstruction fidelity. Table~\ref{tab:vae} compares LPM-VAE with representative image and video VAEs on a challenging internal benchmark. LPM-VAE achieves the highest PSNR and SSIM~\cite{wang2004image} among the evaluated models, demonstrating its ability to preserve fine-grained structures under latent compression.

For the DiT backbone, we replace GeLU~\cite{hendrycks2016gaussian} in the feed-forward network with SwiGLU~\cite{shazeer2020glu} to improve model expressivity. We further replace LayerNorm~\cite{ba2016layer} with RMSNorm~\cite{zhang2019root} for simpler and more efficient normalization. Spatial positions are parameterized using rotary positional embeddings (RoPE)~\cite{su2024roformer}, which encode relative spatial relationships and support generalization across resolutions. Finally, we train the model using Rectified Flow~\cite{liu2023flow}, which defines a linear probability path and provides a simple and effective objective for generative modeling and efficient sampling.

\begin{table}[tbp]
\centering
\renewcommand{\arraystretch}{1.15}
\caption{VAE reconstruction quality on a challenging internal benchmark. The best result in each column is highlighted.}
\label{tab:vae}
\small
\begin{tabular}{l S[table-format=2.2] S[table-format=1.3]}
\toprule
\textbf{Method} & \textbf{PSNR $\uparrow$} & \textbf{SSIM $\uparrow$} \\
\midrule
SD2.1-VAE~\cite{rombach2022highresolution} & 22.56 & 0.467 \\
SDXL-VAE~\cite{podell2023sdxl}     & 22.48 & 0.484 \\
SD3-VAE~\cite{esser2024scaling}    & 25.78 & 0.726 \\
Cosmos-VAE~\cite{reda2024cosmos}   & 27.28 & 0.774 \\
VideoVAE+~\cite{xing2025videovae+} & 25.03 & 0.678 \\
Wan2.1-VAE~\cite{wan2025wan}       & 26.01 & 0.719 \\
\rowcolor{highlightrow}
LPM-VAE & \bfseries 27.66 & \bfseries 0.807 \\
\bottomrule
\end{tabular}
\renewcommand{\arraystretch}{1}
\end{table}

\paragraph{\textbf{Diffusion-Based Image Restoration.}}
Given an HQ image \(I_{\mathrm{HQ}}\) and its corresponding LQ observation \(I_{\mathrm{LQ}}\), generative image restoration aims to model the conditional distribution
\(p_\theta(I_{\mathrm{HQ}}\mid I_{\mathrm{LQ}})\).
We encode the image pair into the latent space using the encoder
\(\mathcal{E}\) of LPM-VAE:
\(z_{\mathrm{HQ}}=\mathcal{E}(I_{\mathrm{HQ}})\) and
\(z_{\mathrm{LQ}}=\mathcal{E}(I_{\mathrm{LQ}})\).

LPM-Image is trained with a conditional flow-matching objective. Let
\(\epsilon\sim\mathcal{N}(0,\mathbf{I})\) denote Gaussian noise and
\(\tau\sim\mathcal{U}[0,1]\) denote the flow timestep. Following
Rectified Flow, we define a linear probability path from the noise
distribution to the HQ latent distribution $z_\tau = (1-\tau)\epsilon+\tau z_{\mathrm{HQ}}$.
Conditioned on the LQ latent \(z_{\mathrm{LQ}}\), the DiT backbone
\(v_\theta\) is optimized to predict this velocity, where \(v^\star\) denotes the target velocity field along the probability
path:
\begin{equation}
    \mathcal{L}_{\mathrm{FM}}
    =
    \mathbb{E}_{\substack{
        \epsilon\sim\mathcal{N}(0,\mathbf{I}),\,
        \tau\sim\mathcal{U}[0,1]
    }}
    \left[
        \left\|
        v_\theta
        \left(
            z_\tau,
            z_{\mathrm{LQ}},
            \tau
        \right)
        -
        v^\star
        \right\|_2^2
    \right].
    \label{eq:flow_matching}
\end{equation}

An important architectural choice is how to inject the LQ condition into
the generative backbone. ControlNet-based methods~\cite{zhang2023adding,
lin2025tasr, qu2024xpsr, shi2025self} introduce a dedicated conditioning
branch, providing strong structural control at the cost of additional
parameters, memory, and training overhead. Attention-based
conditioning~\cite{yang2024pixel, wu2024seesr, li2022srdiff} enables
flexible interactions between conditional and generative features, but
global feature matching may weaken precise local correspondence when
pixel-level alignment is required.

LPM-Image instead adopts a direct and parameter-efficient conditioning
mechanism. Specifically, we spatially align \(z_{\mathrm{LQ}}\) with
\(z_\tau\), concatenate them along the channel dimension, and feed the
resulting representation to the DiT backbone. This design requires no
auxiliary conditioning branch and only modifies the input projection
layer to accommodate the additional channels. More importantly, it
preserves explicit spatial correspondence between the degraded
observation and the restoration target, allowing the model to learn the
LQ-to-HQ mapping end to end. This simple conditioning mechanism provides
an effective balance among reconstruction fidelity, detail generation,
and computational efficiency.

\subsubsection{Training}

LPM-Image is trained progressively in three phases: large-scale pre-training, high-quality fine-tuning, and fidelity refinement. These phases respectively establish broad restoration capabilities, shift the model toward a higher-quality output distribution, and strengthen its faithfulness to input structures.

\paragraph{\textbf{Large-Scale Pre-Training.}}
We first train LPM-Image on the conditional LQ-to-HQ restoration task at a resolution of \(256\times256\). This phase uses a billion-scale training corpus and focuses on learning both the distribution of natural images and the correspondence between diverse degradations and their clean counterparts. Large-scale training exposes the model to broad semantic and visual variations, including scene structures, human and facial characteristics, object layouts, materials, textures, and photographic styles. The resulting knowledge forms a general-purpose visual prior that enables LPM-Image to recover plausible structures and fine details even from severely degraded inputs.

\paragraph{\textbf{High-Quality Fine-Tuning.}}
Large-scale pre-training provides broad coverage, but the learned output distribution is inevitably influenced by the long-tailed quality variation in the pre-training corpus. Consequently, the model may not consistently favor outputs with exceptional perceptual fidelity and rich fine-grained detail. We address this issue by fine-tuning LPM-Image on a curated subset of approximately \(48{,}000\) high-quality images. These samples are selected using substantially stricter criteria for technical quality, texture richness, color reproduction, dynamic range, and aesthetic composition. Fine-tuning on this concentrated high-quality distribution steers the model toward cleaner structures, more natural textures, and higher overall perceptual quality while retaining the generalization capability acquired during large-scale pre-training.

\paragraph{\textbf{Fidelity Refinement.}}
Although generative priors improve perceptual quality, they may alter content that should remain faithful to the LQ observation. This issue is particularly pronounced for spatially sensitive, high-frequency structures, such as text, small facial features, clothing patterns, and fine vegetation, where minor generative deviations can produce incorrect characters, distorted identities, or inconsistent local patterns. Restoration therefore requires not only perceptually plausible details, but also strong structural and pixel-level faithfulness to the input.

To strengthen this property, we construct a dedicated refinement dataset containing text and fine-texture regions with precise region-level annotations. During degradation synthesis, annotated regions probabilistically bypass the degradation operator. Given an HQ image \(I_{\mathrm{HQ}}\), a degradation operator \(\mathcal{D}_{\mathrm{deg}}\), and a binary region mask \(M\), the corresponding LQ input is constructed as
\begin{equation}
    I_{\mathrm{LQ}}
    =
    M\odot I_{\mathrm{HQ}}
    +
    (1-M)\odot
    \mathcal{D}_{\mathrm{deg}}(I_{\mathrm{HQ}}),
    \label{eq:fidelity_refinement}
\end{equation}
This construction introduces region-level HQ-to-HQ identity supervision into the standard LQ-to-HQ restoration task, explicitly teaching the model to preserve already-correct structures rather than regenerate them unnecessarily. We further augment the refinement data with targeted synthesis procedures covering challenging text and fine-detail patterns. Together, these strategies reduce structural hallucination and improve detail preservation without sacrificing the perceptual gains obtained in the preceding stages.

\subsection{Video-Based Model}

Applying LPM-Image independently to individual video frames preserves rich spatial details but often produces temporal artifacts, including flickering, texture inconsistency, and appearance drift. These artifacts arise because frame-wise restoration does not explicitly model temporal dependencies and therefore cannot coordinate generated details across frames. To address this limitation, we extend LPM-Image into \textbf{LPM-Video}, which introduces efficient temporal modeling while retaining the spatial restoration prior learned by LPM-Image. This section presents its architecture, training strategy, and long-video inference procedure.

\begin{figure}[tbp]
    \centering
    \includegraphics[width=0.95\linewidth]{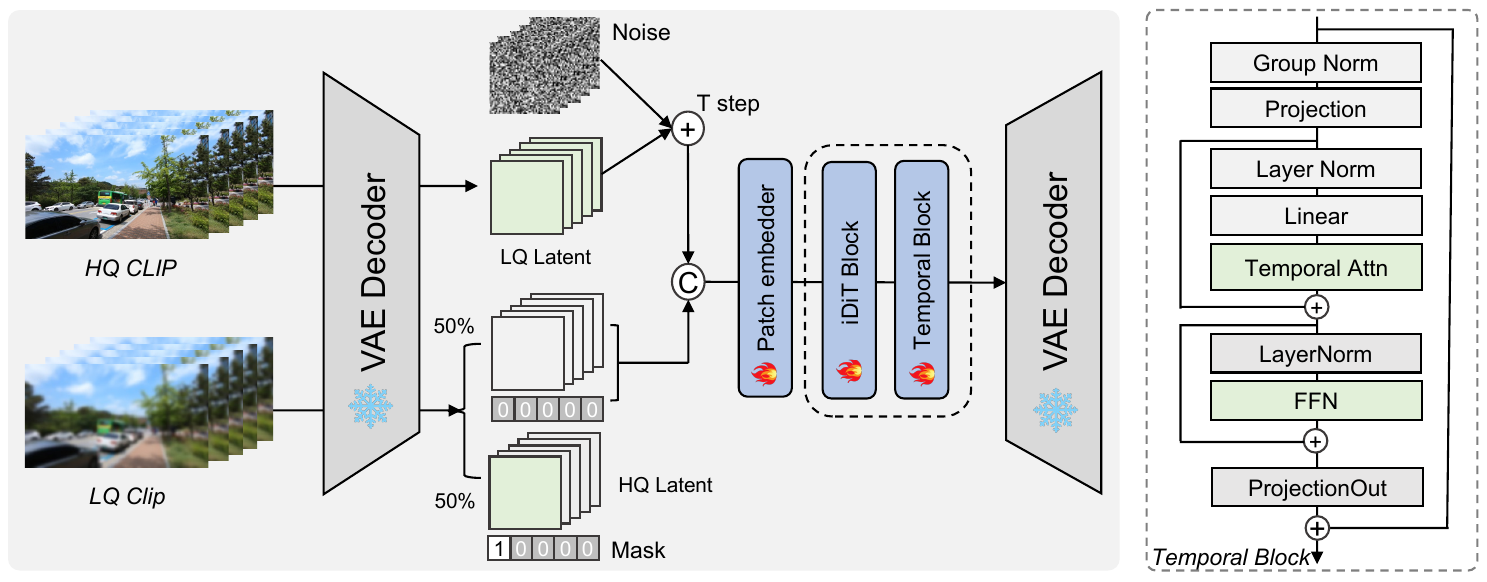}
    \caption{Overview of LPM-Video and its temporal block. LPM-Video augments the spatial DiT blocks of LPM-Image with factorized temporal-attention blocks and employs mask-guided reference conditioning to propagate restored information across clips.}
    \label{fig:video_pipeline}
\end{figure}

\subsubsection{Architecture}

\paragraph{\textbf{Factorized Spatiotemporal Modeling.}}
To introduce temporal modeling without compromising the spatial restoration capability of LPM-Image, we insert a temporal block after each spatial DiT block, following the factorized design of AnimateDiff~\cite{guo2023animatediff}. Each spatial DiT block applies 2D self-attention independently within individual frames, whereas each temporal block applies 1D self-attention across frames at every spatial-token location. Compared with joint attention over all spatiotemporal tokens, this factorization substantially reduces computational and memory costs while enabling effective cross-frame information exchange.

For a batch of \(B\) clips, each containing \(T\) frames and \(N\) spatial tokens per frame, the spatial DiT blocks operate on a representation of shape \((BT)\times N\times D\), where \(D\) denotes the hidden dimension. Before temporal attention, the representation is rearranged to \((BN)\times T\times D\), such that tokens at the same spatial location form a temporal sequence. The temporal block performs self-attention along this sequence, after which the representation is restored to its spatial layout and passed to the next spatial DiT block.

This factorized 2D+1D architecture directly inherits the spatial restoration prior and fine-detail generation capability of LPM-Image. It also retains the frame-wise 2D VAE, avoiding the additional training and reconstruction costs associated with a dedicated 3D-VAE.

\paragraph{\textbf{Position-Free Temporal Attention.}}
LPM-Video does not use explicit temporal positional embeddings, a design we refer to as \textbf{NoPE}. Conventional fixed or learned temporal embeddings associate the model with the temporal windows encountered during training and may hinder extrapolation to longer sequences. By removing absolute temporal indices, NoPE makes the temporal operator independent of a predefined sequence length and allows the same model to accept variable-length inputs. Temporal interactions are instead established through content-dependent attention over the motion and appearance cues present in the LQ sequence.

NoPE removes the architectural restriction imposed by a fixed temporal embedding range, but processing an entire long video in a single forward pass remains constrained by computation and memory. In practice, we therefore restore long videos clip by clip and explicitly propagate high-quality references across adjacent clips.

\paragraph{\textbf{Mask-Guided Cross-Clip Conditioning.}}
Independent clip-wise restoration can introduce visible discontinuities at clip boundaries because fine textures and local appearances may vary across successive windows. We address this issue with \textit{mask-guidance training}, which exposes the model to the recursive conditioning pattern used during long-video inference. Specifically, the last restored frame of the preceding clip is reused as a high-quality temporal anchor for the current clip, while the remaining frames are conditioned on their corresponding LQ observations.

The conditional sequence and its binary mask are provided to the DiT backbone together with the noisy latent. In implementation, the frame-level mask is spatially broadcast and concatenated with the model input along the channel dimension. This explicit indicator allows the model to distinguish restored references from degraded observations and to treat the former as high-quality temporal anchors. By exposing this recursive dependency during training, mask guidance reduces the discrepancy between training and long-video inference.

Combined with the spatial prior inherited from LPM-Image and factorized temporal attention, mask-guided conditioning improves both frame-to-frame and cross-clip consistency while preserving fine-grained spatial details. These components provide the architectural foundation for the arbitrary-length inference strategy introduced below.


\subsubsection{Training}

\begin{figure}[htbp]
    \centering
    \includegraphics[width=0.9\linewidth]{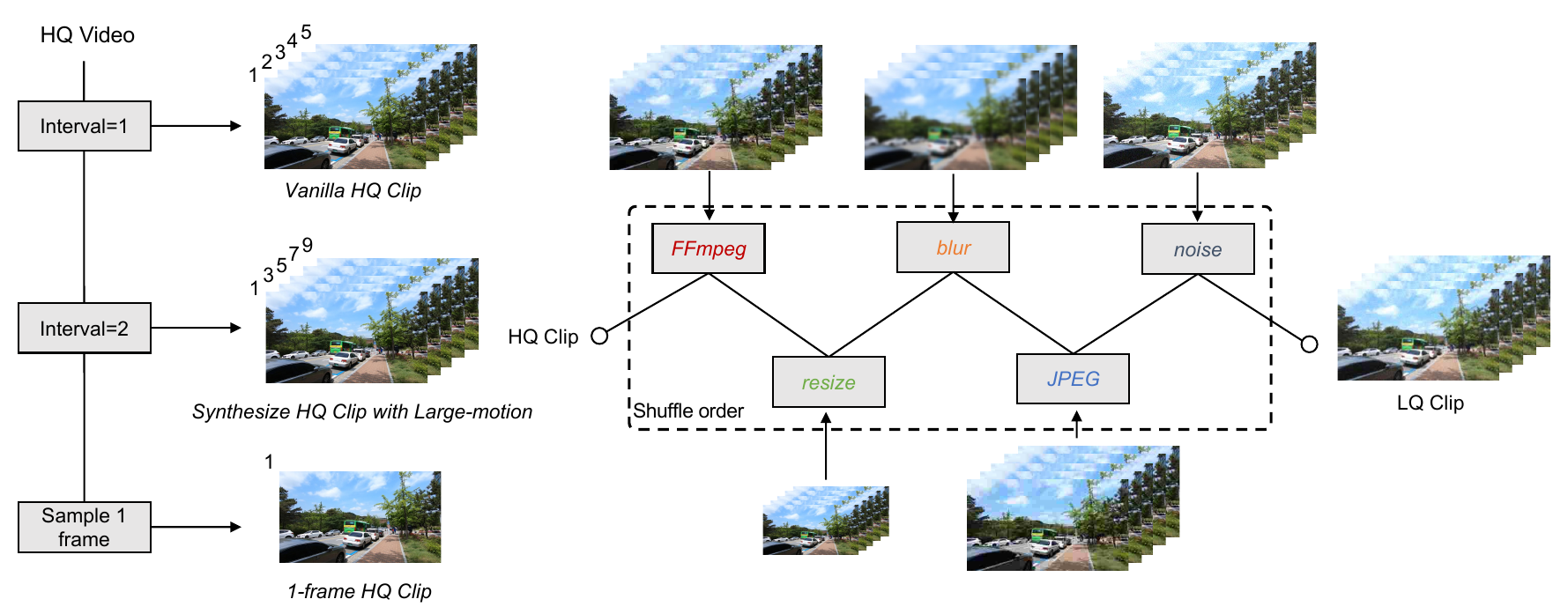}
    \caption{Overview of the degradation and data construction pipeline for LPM-Video training. Image-level degradations are applied consistently within each clip and are followed by FFmpeg-based video compression under multiple bitrate settings.}
    \label{fig:degradation_pipeline}
\end{figure}

\paragraph{\textbf{Temporally Coherent Degradation Pipeline.}}
The degradation pipeline for LPM-Video extends the image degradation pipeline used for LPM-Image. Starting from the BSRGAN degradation model, we additionally introduce FFmpeg-based video compression with multiple bitrate levels to better approximate the compression artifacts encountered in real-world video delivery. The complete pipeline is illustrated in Fig.~\ref{fig:degradation_pipeline}.

A key difference from image restoration is that video degradations are generally temporally correlated. Independently sampling degradation parameters for each frame would introduce artificial temporal variations that rarely occur in real videos and could encourage the model to learn incorrect temporal priors. We therefore share the degradation configuration---including blur, noise, resampling, and compression settings---across all frames within a training clip. This clip-consistent degradation strategy preserves natural temporal continuity while exposing the model to diverse degradation patterns across clips.

\paragraph{\textbf{Progressive Video Training.}}
LPM-Video is trained in two stages: large-scale pre-training and high-quality fine-tuning. The first stage learns general temporal dynamics and intra-clip consistency while preserving the spatial restoration prior of LPM-Image. The second stage further improves perceptual quality and introduces explicit cross-clip consistency for long-video restoration.

\paragraph{\textbf{Large-Scale Pre-Training.}}
We initialize the spatial DiT blocks of LPM-Video from LPM-Image and train the newly introduced temporal blocks jointly with the inherited image backbone. This initialization transfers the spatial restoration and detail-generation capabilities learned by LPM-Image, allowing video training to focus on temporal correspondence and feature aggregation.

Pre-training uses a large-scale mixture of millions of video clips and images. Images are treated as single-frame sequences (\(T=1\)) and sampled together with multi-frame video clips. This mixed-frame strategy mitigates the degradation of spatial generation quality that may otherwise arise when optimizing predominantly for temporal smoothness. At this stage, mask-guidance training is disabled, allowing the model to first learn stable intra-clip temporal dynamics without the additional complexity of recursive cross-clip conditioning.

\paragraph{\textbf{High-Quality Fine-Tuning.}}
We subsequently fine-tune LPM-Video on a curated dataset containing thousands of high-quality video clips. In addition to improving spatial detail and perceptual quality, this stage introduces mask-guidance training to align the training procedure with clip-wise long-video inference. Specifically, consecutive clips are sampled from the same shot, and the last frame associated with the preceding clip is used as the reference for the current clip.

To simulate a previously restored high-quality anchor, we replace the LQ condition at the reference position with its corresponding HQ frame with a probability of \(50\%\) and mark this position using the guidance mask. Otherwise, the model receives the original LQ condition. This stochastic construction prevents excessive dependence on an ideal reference while teaching the model to propagate high-quality appearance and texture information across clip boundaries. We retain mixed-frame training during fine-tuning by probabilistically sampling images as single-frame sequences, thereby preserving spatial generation quality while strengthening both intra-clip and cross-clip temporal consistency.

\paragraph{\textbf{Large-Motion Adaptation.}}
After high-quality fine-tuning, we observe that residual flickering occurs primarily in videos with large inter-frame motion. We attribute this behavior to a systematic imbalance in the training distribution: large motion is often accompanied by motion blur, while our stringent visual-quality filters preferentially remove blurred clips. As a result, clips that simultaneously exhibit high visual quality and large motion are underrepresented in the fine-tuning data.

We address this limitation in two complementary ways. First, we supplement the training corpus with additional real-world clips that contain both high-quality content and large motion. Second, we temporally subsample high-frame-rate or low-motion videos using larger frame intervals, thereby increasing the apparent inter-frame displacement without synthetically introducing motion blur. Fine-tuning on this motion-enhanced data improves temporal stability and detail preservation under large motion.


\subsection{Inference}

\paragraph{\textbf{Limitations of Sliding-Window Inference.}}
Long-video restoration requires temporal consistency over a sequence whose duration may be effectively unbounded. Processing an entire video in a single forward pass provides the model with global temporal context, but is impractical because the computational and memory costs of temporal attention grow rapidly with sequence length. In practice, long videos are therefore commonly divided into overlapping windows and restored sequentially.

Let \(\mathcal{H}_\theta\) denote the LPM-Video restoration operator, \(z_{\mathrm{LQ},1:T}\) the LQ latent sequence, and \(\hat{z}_{1:T}\) the restored sequence. Given a window \(\mathcal{W}_j\) and its overlap \(\mathcal{O}_j\) with the preceding window, conventional sliding-window inference can be written abstractly as
\begin{equation}
    \hat{z}_{\mathcal{W}_j\setminus\mathcal{O}_j}
    =
    \mathcal{H}_\theta
    \left(
        z_{\mathrm{LQ},\mathcal{W}_j\setminus\mathcal{O}_j};
        \hat{z}_{\mathcal{O}_j}
    \right),
    \label{eq:sliding_window}
\end{equation}
where the restored overlapping frames \(\hat{z}_{\mathcal{O}_j}\) serve as temporal references for the current window. Although this mechanism improves local continuity, each window is conditioned only on the output of its immediate predecessor. Consequently, small reconstruction discrepancies can propagate through the inference chain and gradually accumulate over long sequences. The resulting drift is particularly visible in static or low-motion regions, where subtle variations in texture or color are readily perceived as flickering.

Keyframe-based inference partially alleviates this problem by restoring a sparse set of keyframes before processing the intermediate frames. However, a single-level keyframe strategy can still become unstable when adjacent keyframes are temporally distant, when shots vary substantially in duration, or when complex scene transitions are present.

\begin{figure}[tbp]
    \centering
    \includegraphics[width=0.9\linewidth]{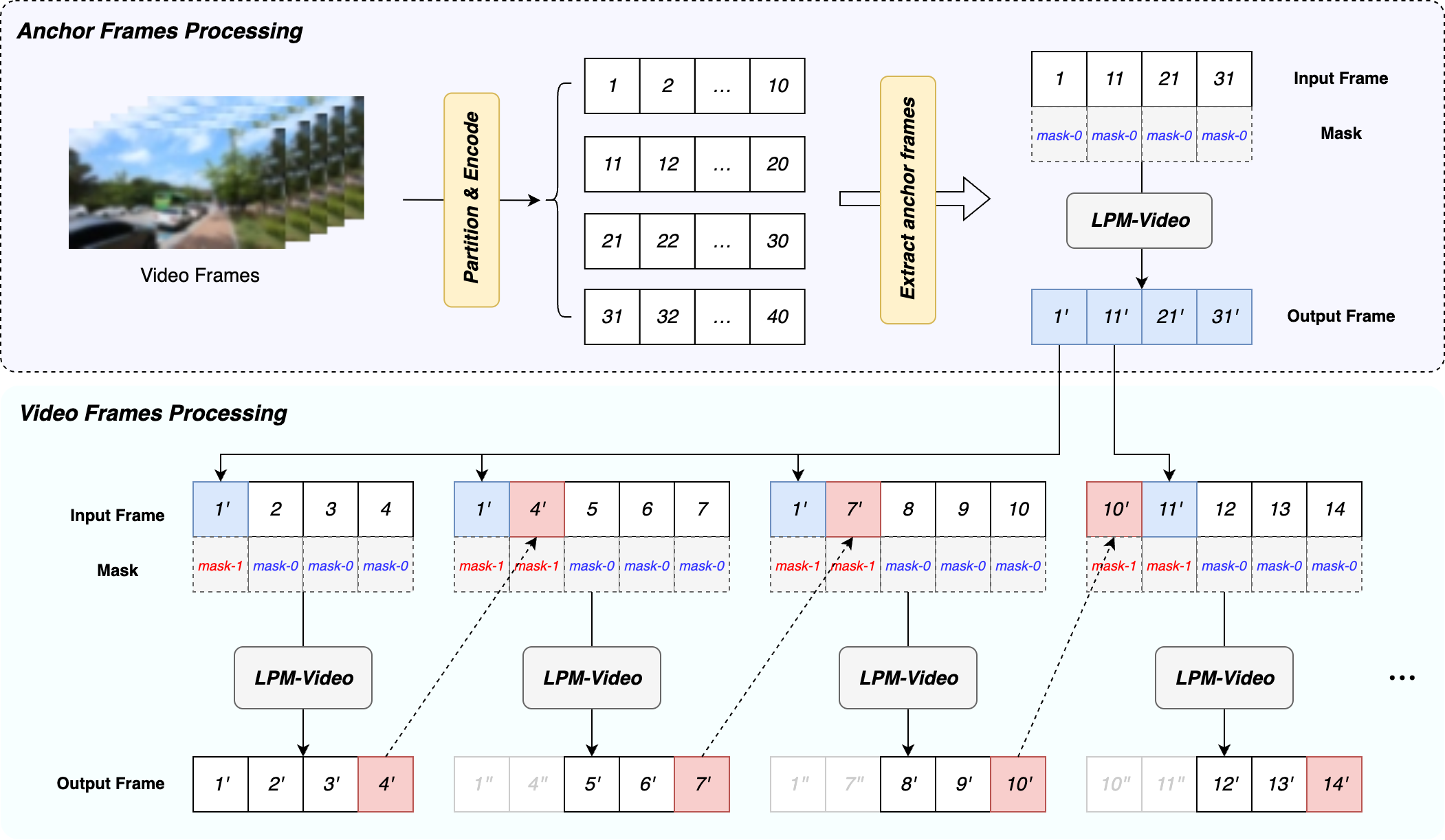}
    \caption{Temporal-pyramid inference. Restored shot-level keyframes provide long-range guidance, while overlapping frames maintain short-range continuity. Long shots are handled recursively.}
    \label{fig:pyramid_infer}
\end{figure}

\paragraph{\textbf{Temporal-Pyramid Inference.}}
To capture both long-range consistency and short-range continuity, we introduce a {temporal-pyramid inference} strategy, as illustrated in Fig.~\ref{fig:pyramid_infer}. The central idea is to construct temporally coherent anchors at a coarse level and use them to guide restoration at progressively finer temporal scales.

We first partition the input video into \(N_s\) shots using scene-cut detection. Let \(a_s\) denote the index of the first frame of the \(s\)-th shot. The corresponding shot-level keyframes are collected into a sparse sequence and restored jointly:
\begin{equation}
    \left\{
        \hat{z}_{a_s}
    \right\}_{s=1}^{N_s}
    =
    \mathcal{H}_\theta
    \left(
        \left\{
            z_{\mathrm{LQ},a_s}
        \right\}_{s=1}^{N_s};
        \emptyset
    \right).
    \label{eq:storyboard}
\end{equation}
The restored keyframes form a high-quality \emph{storyboard} that provides a set of stable, shot-level anchors for the full video. Joint restoration at this sparse temporal scale encourages coherent appearance and restoration characteristics across the video while preserving the semantic discontinuities introduced by scene cuts.

Each shot is subsequently restored using sliding windows. For a window \(\mathcal{W}_{s,j}\) in shot \(s\), LPM-Video is conditioned on two complementary references: the fixed shot-level anchor \(\hat{z}_{a_s}\), which provides long-range guidance, and the restored overlap \(\hat{z}_{\mathcal{O}_{s,j}}\), which maintains short-range continuity:
\begin{equation}
    \hat{z}_{\mathcal{W}_{s,j}\setminus\mathcal{O}_{s,j}}
    =
    \mathcal{H}_\theta
    \left(
        z_{\mathrm{LQ},
        \mathcal{W}_{s,j}\setminus\mathcal{O}_{s,j}};
        \hat{z}_{a_s},
        \hat{z}_{\mathcal{O}_{s,j}}
    \right).
    \label{eq:pyramid_inference}
\end{equation}
The reference type is indicated by the guidance mask introduced during training. Unlike conventional sliding-window inference, which relies exclusively on recursively generated local references, every window is connected to a stable shot-level anchor. This shortens the effective dependency chain and prevents local reconstruction errors from freely propagating throughout the shot.

Because LPM-Video uses position-free temporal attention, reference frames can be incorporated into inference windows without introducing temporal indices outside the range observed during training. The guidance mask explicitly identifies their roles, allowing the same model to process varying numbers of frames and references.

\paragraph{\textbf{Recursive Extension to Arbitrary-Length Videos.}}
A single shot may itself contain more frames than can be covered reliably by one level of windowed inference. In this case, we recursively apply the same construction within the shot: sparse intermediate anchors are first restored under the guidance of the shot-level keyframe, and the intervals between adjacent anchors are then restored at the next finer level. Recursion terminates when each segment can be processed within the supported temporal window.

This hierarchical construction extends LPM-Video to videos of arbitrary duration without requiring a single full-sequence forward pass. Since only sparse anchors are processed at the upper levels of the pyramid, the additional computation is small relative to frame-level restoration. More importantly, the combination of coarse long-range anchors and fine local references substantially reduces error accumulation and appearance drift, enabling stable restoration over long videos and complex shot structures.

\subsection{Acceleration}

Accelerating LPM poses different challenges than accelerating T2V models. T2V acceleration chiefly preserves semantic consistency across frames rather than pixel-level fidelity, whereas LPM must recover fine-grained textures and high-frequency detail from low-quality inputs without introducing pixel-level discrepancies—a substantially more constrained optimization objective.

T2V acceleration typically retains around 5 sampling steps (e.g., SDXL-Turbo), but LPM must process long-form UGC spanning minutes to hours, which demands higher throughput; our target is 1- or 2-step sampling. At such low step counts, the ``denoising'' and ``generation'' phases of the diffusion process collapse into a single forward pass~\cite{lee2025truncated}, which can compromise the model's ability to reconcile restoration with generation and lead to high-frequency collapse or texture blurring.

\subsubsection{Training-Aware Acceleration}
To overcome the efficiency bottleneck of multi-step diffusion sampling, we adopt a training-aware acceleration approach based on Consistency Models (CM)~\cite{song2023consistency}, distilling the 25-step Rectified Flow (RF) trajectory into a single-step student model.

\begin{figure}[tbp]
    \centering
    \includegraphics[
        width=0.8\linewidth,
        clip
    ]{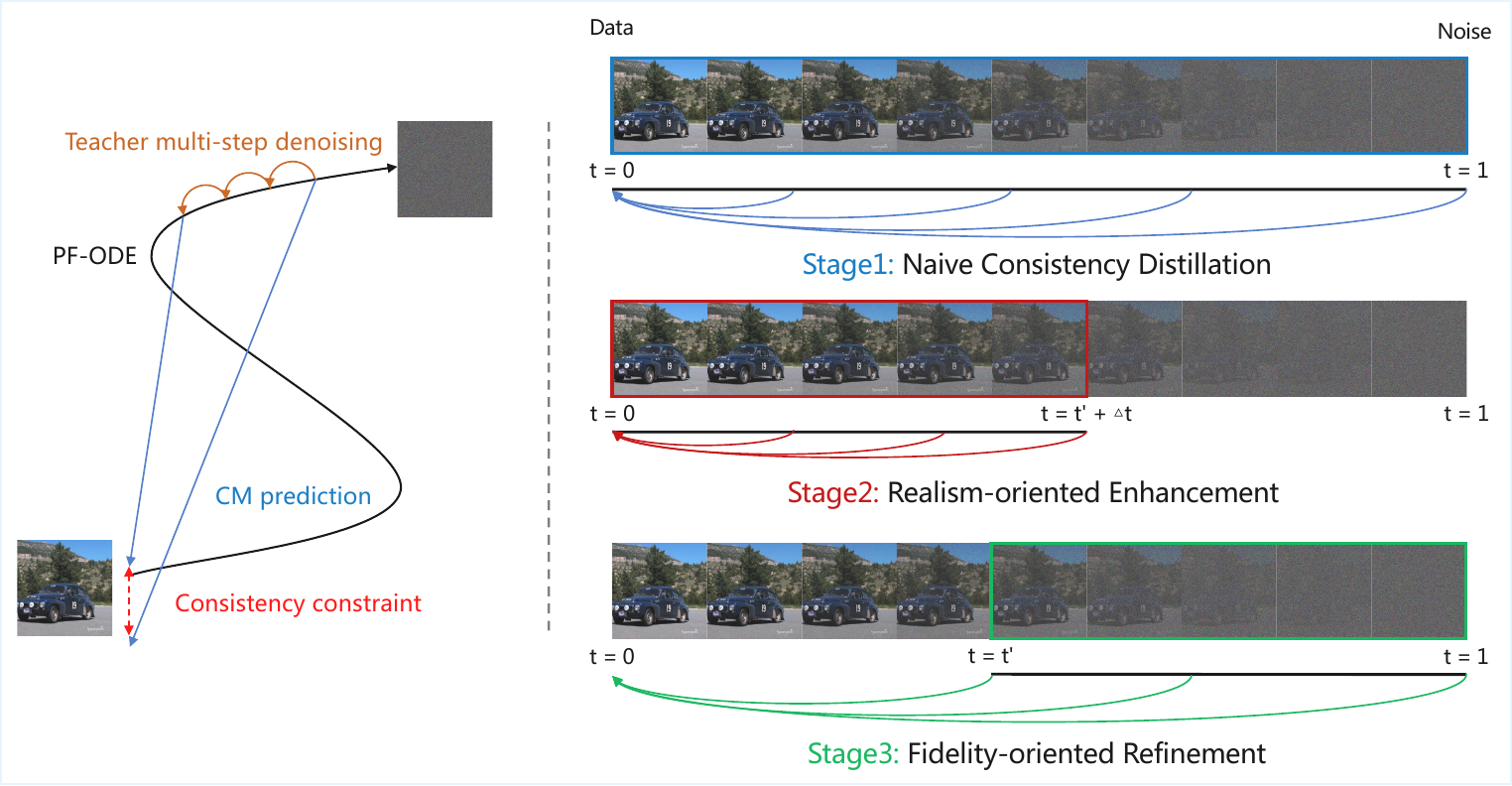}
    \caption{Trajectory-based Distillation in LPM: (left) multi-step integration, (right) 3-stage TCM.}
    \label{fig:CM-figure1}
\end{figure}

\begin{itemize}

    \item \textbf{Velocity Reparameterization}: Consistency models enforce coherence in state space, which is incompatible with LPM's velocity-based sampling. We bridge this gap by reparameterizing the predicted velocity as an explicit state via $x_{t_n}=x_{t_{n+1}}+(t_{n+1}-t_n)v$, aligning the student's output space with the teacher's velocity predictions. This allows effective reuse of pre-trained parameters while preserving the consistency objective across adjacent timesteps.

    \item \textbf{Multi-step Integration}: To mitigate the optimization instability commonly observed in consistency-model training, we adopt the skipping-step strategy from Latent Consistency Models (LCM)~\cite{luo2023latent}, which distributes supervision across multiple intermediate timesteps rather than enforcing the one-step objective directly. This yields more reliable gradients and faster convergence during early-stage distillation, and lets the teacher sample more densely along the trajectory, improving distillation fidelity.

    \item \textbf{Three-Stage Truncated CM Training}: To resolve the conflict between ``denoising'' and ``generation'' in one-step inference, we propose a \textit{three-stage truncated consistency model (3-stage TCM)} training strategy. As shown in Fig.~\ref{fig:CM-figure1}, unlike the original two-stage TCM~\cite{lee2025truncated}, we insert an additional intermediate stage between Stage~1 and Stage~2 that focuses exclusively on timesteps near the clean image, targeting fine-grained detail generation. This three-stage pipeline (naive consistency distillation $\rightarrow$ realism-oriented enhancement $\rightarrow$ fidelity-oriented refinement) achieves a better balance between restoration fidelity and generative detail quality than the two-stage baseline.

\end{itemize}

\subsubsection{Training-Free Acceleration}
We complement training-aware distillation with training-free optimizations for practical deployment:
\begin{itemize}
    \item \textbf{DiT Optimization}: We apply FP8 quantization to the DiT backbone, tuned for contemporary GPU architectures, and use SageAttention~\cite{zhang2025sageattention} for the precision-sensitive attention layers, reducing computational complexity for high-resolution, long-duration video.
    \item \textbf{VAE Optimization}: The VAE encoder/decoder is quantized to INT8. We implement tiled VAE inference, processing high-resolution frames tile-by-tile, which reduces peak GPU memory and improves batched-inference throughput.
    \item \textbf{Deployment SDK}: The full inference pipeline is deployed via TensorRT-LLM~\cite{tensorrt_llm}. Operator fusion and custom CUDA kernels further reduce latency, yielding an optimized end-to-end video processing pipeline.
\end{itemize}

Layering training-free optimization on top of training-aware distillation, these techniques together yield a 39.5$\times$ end-to-end speedup over the original 25-step pipeline without sacrificing visual quality (PSNR $>$ 43\,dB, SSIM $>$ 0.99, KVQ difference $<$ 0.01 relative to the unaccelerated model), enabling large-scale deployment in practice. Table~\ref{tab:acceleration} summarizes the cumulative contribution of each stage. When visual fidelity takes precedence over latency, an optional 2-step sampling configuration further improves high-frequency detail reconstruction while retaining most of the speedup over standard multi-step diffusion.

\begin{table}[htbp]
\centering
\renewcommand{\arraystretch}{1.15}
\caption{Cumulative end-to-end speedup as training-aware and training-free acceleration are layered on top of the 25-step baseline.}
\label{tab:acceleration}
\small
\begin{tabular}{c S[table-format=2.1] c}
\toprule
\textbf{Stage} & \textbf{Cumulative Speedup $\uparrow$} & \textbf{Approach} \\
\midrule
+ Training-aware & \cellcolor{highlightrow}12.4 $\times$ & LCM, 3-stage TCM \\
+ Training-free  & \cellcolor{highlightrow}39.5 $\times$ & FP8/INT8 quantization, TensorRT-LLM deployment \\
\bottomrule
\end{tabular}
\renewcommand{\arraystretch}{1}
\end{table}

\section{Experiments}

\subsection{Benchmarks and Metrics}

We evaluate LPM using complementary full-reference and no-reference image-quality metrics. For benchmarks with ground-truth HQ targets, we report peak signal-to-noise ratio (PSNR) to measure pixel-level reconstruction fidelity. We additionally report MUSIQ~\cite{ke2021musiq}, CLIP-IQA~\cite{wang2023exploring}, and KVQ~\cite{lu2024kvq}, which assess perceptual quality without requiring reference images. For image restoration, we evaluate LPM-Image on two public benchmarks: RealSR~\cite{cai2019toward} and RealSet~\cite{yue2023resshift}. 
For video restoration, we evaluate LPM-Video on four public benchmarks: VideoLQ~\cite{chan2022investigating}, YouHQ~\cite{zhou2024upscale}, REDS~\cite{nah2019ntire}, and SPMCS~\cite{yi2019progressive}. To assess robustness under a broader range of in-the-wild content and degradation patterns, we further construct \textbf{LPM-Benchmark}, which contains 192 videos covering diverse semantic categories, including humans, animals, natural scenes, and architecture. Frames sampled from these videos form a corresponding image-restoration benchmark. All methods are evaluated using the same input content, output resolutions, and metric implementations.

\subsection{Quantitative Comparisons}

Table~\ref{tab:image-result-1} compares LPM-Image with representative diffusion-based restoration methods, including StableSR~\cite{wang2024exploiting}, SeeSR~\cite{wu2024seesr}, and DiT-SR~\cite{cheng2025effective}, on the public RealSR and RealSet benchmarks. LPM-Image achieves the highest MUSIQ and KVQ scores on both datasets. On RealSR, SeeSR obtains a slightly higher PSNR (\(26.35\) vs.\ \(25.97\) dB), whereas LPM-Image substantially improves KVQ from \(3.23\) to \(3.98\). This result reflects the perception--distortion trade-off commonly observed in generative restoration: LPM-Image maintains competitive pixel-level fidelity while producing more natural and perceptually convincing details.

\begin{table}[htbp]
\centering
\renewcommand{\arraystretch}{1.15}
\caption{Quantitative comparison for image super-resolution on the RealSR and RealSet benchmarks.}
\label{tab:image-result-1}
\small
\begin{tabular}{c l
S[table-format=2.2] S[table-format=2.2] S[table-format=1.4] S[table-format=1.2]}
\toprule
\textbf{Dataset} & \textbf{Method} & \textbf{PSNR $\uparrow$} & \textbf{MUSIQ $\uparrow$} & \textbf{CLIP-IQA $\uparrow$} & \textbf{KVQ $\uparrow$} \\
\midrule
\multirow{4}{*}{RealSR~\cite{cai2019toward}}
 & StableSR~\cite{wang2024exploiting} & 24.05 & 57.51 & 0.5541 & 2.86 \\
 & SeeSR~\cite{wu2024seesr}     & \bfseries 26.35 & 62.81 & 0.5591 & 3.23 \\
 & DiT-SR~\cite{cheng2025effective}    & 24.21 & 64.33 & 0.5909 & 3.47 \\
 & \cellcolor{highlightrow}LPM-Image & \cellcolor{highlightrow}25.97 & \cellcolor{highlightrow}\bfseries 64.68 & \cellcolor{highlightrow}\bfseries 0.6356 & \cellcolor{highlightrow}\bfseries 3.98 \\
\midrule
\multirow{4}{*}{RealSet~\cite{yue2023resshift}}
 & StableSR~\cite{wang2024exploiting} & {--} & 50.00 & 0.6499 & 3.28 \\
 & SeeSR~\cite{wu2024seesr}     & {--} & 63.73 & 0.5858 & 3.80 \\
 & DiT-SR~\cite{cheng2025effective}    & {--} & 65.95 & \bfseries 0.6765 & 3.64 \\
 & \cellcolor{highlightrow}LPM-Image & \cellcolor{highlightrow}{--} & \cellcolor{highlightrow}\bfseries 66.67 & \cellcolor{highlightrow}0.6364 & \cellcolor{highlightrow}\bfseries 3.95 \\
\bottomrule
\end{tabular}
\renewcommand{\arraystretch}{1}
\end{table}

To further evaluate generalization, we compare LPM-Image with 13 competitive methods on LPM-Benchmark. As shown in Table~\ref{tab:image-result-2}, LPM-Image achieves the best result across all three metrics, outperforming recent strong baselines. The improvement is particularly pronounced in KVQ, where LPM-Image reaches \(4.046\), compared with \(3.634\) for the strongest competing methods. These consistent gains suggest that LPM-Image generalizes effectively across the diverse content categories and real-world degradation patterns represented in LPM-Benchmark.

\begin{table}[tbp]
\centering
\renewcommand{\arraystretch}{1.15}
\caption{Quantitative comparison for image super-resolution on LPM-Benchmark.}
\label{tab:image-result-2}
\small
\begin{tabular}{l
S[table-format=2.2] S[table-format=1.4] S[table-format=1.3] |
l S[table-format=2.2] S[table-format=1.4] S[table-format=1.3]}
\toprule
\textbf{Method} & \textbf{MUSIQ $\uparrow$} & \textbf{CLIP-IQA $\uparrow$} & \textbf{KVQ $\uparrow$} & \textbf{Method} & \textbf{MUSIQ $\uparrow$} & \textbf{CLIP-IQA $\uparrow$} & \textbf{KVQ $\uparrow$} \\
\midrule
StableSR~\cite{wang2024exploiting} & 70.44 & 0.6289 & 3.542 & CCSR~\cite{sun2023improving}     & 71.90 & 0.7043 & 3.467 \\
SeeSR~\cite{wu2024seesr}           & 69.16 & 0.6154 & 3.450 & PiSA-SR~\cite{sun2025pixel}      & 72.85 & 0.7230 & 3.634 \\
ResShift~\cite{yue2023resshift}    & 62.69 & 0.5515 & 3.191 & AdcSR~\cite{chen2025adversarial} & 69.96 & 0.6667 & 3.475 \\
DiT-SR~\cite{cheng2025effective}   & 64.97 & 0.5465 & 3.119 & SUPIR~\cite{yu2024scaling}       & 58.39 & 0.4445 & 3.216 \\
S3Diff~\cite{zhang2024degradation} & 62.65 & 0.5405 & 3.239 & TSD-SR~\cite{dong2025tsd}        & 72.43 & 0.7109 & 3.634 \\
InvSR~\cite{yue2025arbitrary}      & 67.23 & 0.6215 & 3.261 & VARSR~\cite{qu2025visual}        & 71.76 & 0.7000 & 3.479 \\
XPSR~\cite{qu2024xpsr}             & 63.64 & 0.5852 & 3.262 & \cellcolor{highlightrow}LPM-Image & \cellcolor{highlightrow}\bfseries 73.32 & \cellcolor{highlightrow}\bfseries 0.7423 & \cellcolor{highlightrow}\bfseries 4.046 \\
\bottomrule
\end{tabular}
\renewcommand{\arraystretch}{1}
\end{table}

Table~\ref{tab:video-result-all} compares LPM-Video against SeedVR2~\cite{wang2025seedvr2}, FlashVSR~\cite{zhuang2025flashvsr}, and Vivid-VR~\cite{bai2025vivid} on five datasets: SPMC, REDS, YouHQ, VideoLQ, and LPM-Benchmark. LPM-Video achieves the best score on every metric across all five. The margin is most pronounced on VideoLQ, where LPM-Video improves MUSIQ from 62.76 (FlashVSR) to 70.46 and KVQ from 2.94 (Vivid-VR) to 3.17. These results indicate that LPM-Video improves not only pixel-level fidelity but also perceptual quality and spatiotemporal consistency, across both image and video restoration settings.

\begin{table}[tbp]
\centering
\renewcommand{\arraystretch}{1.15}
\caption{Quantitative comparison of video restoration methods on public benchmarks and LPM-Benchmark.}
\label{tab:video-result-all}
\small
\begin{tabular}{c l
S[table-format=2.2]
S[table-format=2.2]
S[table-format=1.4]
S[table-format=1.2]}
\toprule
\textbf{Dataset} &
\textbf{Method} &
\textbf{PSNR $\uparrow$} &
\textbf{MUSIQ $\uparrow$} &
\textbf{CLIP-IQA $\uparrow$} &
\textbf{KVQ $\uparrow$} \\
\midrule
\multirow{4}{*}{SPMCS~\cite{yi2019progressive}}
 & SeedVR2~\cite{wang2025seedvr2}
 & 20.46 & 66.87 & 0.5307 & 2.24 \\
 & FlashVSR~\cite{zhuang2025flashvsr}
 & 21.46 & 71.05 & 0.5792 & 3.31 \\
 & Vivid-VR~\cite{bai2025vivid}
 & 21.73 & 70.03 & 0.4832 & 3.12 \\
 & \cellcolor{highlightrow}LPM-Video
 & \cellcolor{highlightrow}\bfseries 22.47
 & \cellcolor{highlightrow}\bfseries 72.87
 & \cellcolor{highlightrow}\bfseries 0.6434
 & \cellcolor{highlightrow}\bfseries 3.89 \\
\midrule
\multirow{4}{*}{REDS~\cite{nah2019ntire}}
 & SeedVR2~\cite{wang2025seedvr2}
 & 21.45 & 61.83 & 0.3695 & 2.37 \\
 & FlashVSR~\cite{zhuang2025flashvsr}
 & 20.76 & 68.97 & 0.4661 & 3.24 \\
 & Vivid-VR~\cite{bai2025vivid}
 & 20.45 & 69.07 & 0.4764 & 3.43 \\
 & \cellcolor{highlightrow}LPM-Video
 & \cellcolor{highlightrow}\bfseries 22.21
 & \cellcolor{highlightrow}\bfseries 71.70
 & \cellcolor{highlightrow}\bfseries 0.6241
 & \cellcolor{highlightrow}\bfseries 3.88 \\
\midrule
\multirow{4}{*}{YouHQ~\cite{zhou2024upscale}}
 & SeedVR2~\cite{wang2025seedvr2}
 & 22.01 & 62.31 & 0.4909 & 2.93 \\
 & FlashVSR~\cite{zhuang2025flashvsr}
 & 21.49 & 69.16 & 0.5873 & 4.02 \\
 & Vivid-VR~\cite{bai2025vivid}
 & 21.31 & 70.55 & 0.4470 & 3.44 \\
 & \cellcolor{highlightrow}LPM-Video
 & \cellcolor{highlightrow}\bfseries 23.55
 & \cellcolor{highlightrow}\bfseries 72.37
 & \cellcolor{highlightrow}\bfseries 0.6575
 & \cellcolor{highlightrow}\bfseries 4.33 \\
\midrule
\multirow{4}{*}{VideoLQ~\cite{chan2022investigating}}
 & SeedVR2~\cite{wang2025seedvr2}
 & {--} & 61.83 & 0.2570 & 2.57 \\
 & FlashVSR~\cite{zhuang2025flashvsr}
 & {--} & 62.76 & 0.4184 & 2.52 \\
 & Vivid-VR~\cite{bai2025vivid}
 & {--} & 58.72 & 0.3948 & 2.94 \\
 & \cellcolor{highlightrow}LPM-Video
 & \cellcolor{highlightrow}{--}
 & \cellcolor{highlightrow}\bfseries 70.46
 & \cellcolor{highlightrow}\bfseries 0.5194
 & \cellcolor{highlightrow}\bfseries 3.17 \\
\midrule
\multirow{4}{*}{LPM-Benchmark}
 & SeedVR2~\cite{wang2025seedvr2}
 & {--} & 56.59 & 0.3763 & 3.01 \\
 & FlashVSR~\cite{zhuang2025flashvsr}
 & {--} & 66.69 & 0.4765 & 3.41 \\
 & Vivid-VR~\cite{bai2025vivid}
 & {--} & 44.42 & 0.3939 & 3.20 \\
 & \cellcolor{highlightrow}LPM-Video
 & \cellcolor{highlightrow}{--}
 & \cellcolor{highlightrow}\bfseries 72.11
 & \cellcolor{highlightrow}\bfseries 0.5954
 & \cellcolor{highlightrow}\bfseries 3.63 \\
\bottomrule
\end{tabular}
\renewcommand{\arraystretch}{1}
\end{table}

\subsection{Qualitative Comparisons}

\begin{figure*}[tbp]
    \centering
    \includegraphics[
        width=1.0\linewidth,
    ]{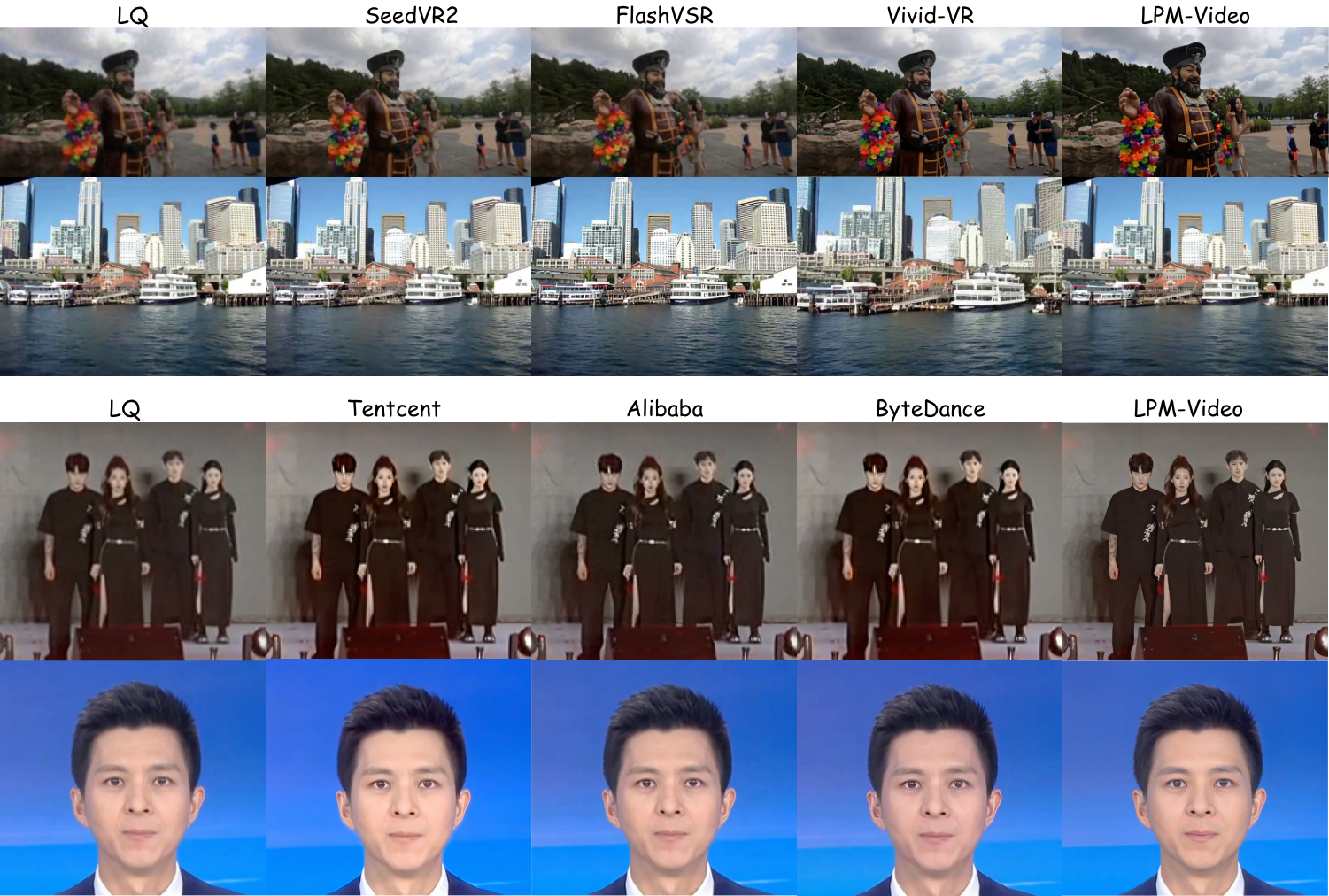}
    \caption{Comparison with open-source methods and external commercial methods.}
    \label{fig:compare1}
\end{figure*}

Figure~\ref{fig:compare1} qualitatively compares LPM-Video with representative open-source methods and commercial video-enhancement services from Tencent~\cite{tencent_mps_av_enhancement}, Alibaba~\cite{aliyun_enhancevidequality_api}, and ByteDance~\cite{volcengine_doc_70279}. For scenes containing complex foreground textures and architectural structures, SeedVR2 and FlashVSR tend to produce blurred boundaries or color bleeding, whereas LPM-Video recovers sharper edges and more coherent fine-scale details. The evaluated commercial services effectively suppress noise and compression artifacts, but their outputs are often over-smoothed, resulting in the loss of subtle facial features and textile patterns. In comparison, LPM-Video better balances artifact removal and detail reconstruction, preserving clearer facial structures, foreground textures, and background content.

The qualitative results further demonstrate the effectiveness of LPM-Video across diverse content types, including natural landscapes, human-centric portraits, architecture, and cinematic scenes. In landscapes such as mountains and bamboo forests, LPM-Video suppresses compression artifacts while recovering fine structures in rock surfaces and foliage. In portrait scenes, it restores facial features and natural skin texture while avoiding the excessively smooth appearance observed in several competing results. LPM-Video also preserves spatially sensitive structures such as embedded text and subtitles, substantially improving their legibility and edge definition. Overall, these examples show that LPM-Video consistently enhances perceptual quality across diverse real-world content while preserving the underlying semantic structures and visual characteristics of the input.

\section{Conclusion and Future Work}

We introduced the \textbf{Large Processing Model (LPM)}, a diffusion-based framework for high-fidelity video restoration under diverse real-world degradations. To our knowledge, LPM is the \textbf{first generative video restoration model deployed at industrial scale}. LPM combines large-scale data engineering with progressive image-to-video training. LPM-Image learns a strong spatial restoration prior, while LPM-Video introduces factorized temporal modeling and mask-guided conditioning to improve temporal consistency. Temporal-pyramid inference further supports stable, clip-wise restoration of videos with arbitrary duration.

In production, LPM serves videos representing approximately \(45\%\) of the total viewing time on Kuaishou and consistently improves key online QoE and user-engagement metrics. When integrated into Kuaishou's encoding and transmission pipeline, it reduces bitrate by more than \(20\%\) at comparable perceptual quality. These results demonstrate the practical value of generative restoration in large-scale video systems.

LPM has two main limitations. First, despite category-aware sampling, the training distribution remains concentrated on common UGC categories, and performance is less consistent on long-tail content such as uncommon outdoor and natural scenes. Second, the perceptual gains are smaller when the input is already of high quality. Future work will focus on broader and more balanced training data, more generalizable restoration priors, and quality-adaptive mechanisms that enhance degraded regions while preserving well-restored content.

\section{Contributors}
Core contributors are listed in alphabetical order by first name.

\textbf{Core Contributors}

\begin{tabularx}{\textwidth}{@{}YYYYYY@{}}
Bichuan Zhu  & Fulin Li     & Jiachao Gong & Jinhua Hao & Kai Zhao     & Kun Yuan \\
Pengcheng Xu & Qiang Wang   & Qiao Mo    & Yanlong Yuan & Yizhen Shao & Yuxiao Hu \\ 
Zixi Tuo \\
\end{tabularx}

\vspace{0.5em}

\textbf{Project Leads}

\begin{tabularx}{\textwidth}{@{}YYYYYY@{}}
Ming Sun & Chao Zhou & Bin Chen & Bin Yu & Bin Wang & \\
\end{tabularx}

\newpage

\clearpage
\newpage
\bibliographystyle{kwaivideo/plainnat}
\bibliography{paper}

\clearpage
\newpage

\end{document}